\documentclass[letterpaper]{article} 
\usepackage{aaai2027}  
\usepackage[hyphens]{url}  
\usepackage{graphicx} 
\usepackage{natbib}  
\usepackage{caption} 
\usepackage{xcolor}

\newcommand{\algstage}[1]{%
  \item[]\hspace{0.5em}%
  {\scriptsize\ttfamily // #1}%
}

\usepackage{algorithm}
\usepackage{algorithmic}
\usepackage{newfloat}
\usepackage{listings}
\DeclareCaptionStyle{ruled}{labelfont=normalfont,labelsep=colon,strut=off} 
\definecolor{promptbg}{RGB}{244,246,251}
\definecolor{promptrule}{RGB}{146,160,190}
\definecolor{tracebg}{RGB}{251,247,237}
\definecolor{tracerule}{RGB}{198,176,136}
\newenvironment{pbox}{%
	\par\addvspace{6pt}\noindent\minipage{\columnwidth}}{%
	\endminipage\par\addvspace{6pt}}
\lstdefinestyle{promptbox}{%
	basicstyle={\scriptsize\ttfamily},
	numbers=none,
	frame=single,framerule=0.5pt,framesep=0pt,
	rulecolor=\color{promptrule},
	backgroundcolor=\color{promptbg},
	framexleftmargin=4pt,framexrightmargin=4pt,
	framextopmargin=4pt,framexbottommargin=4pt,
	xleftmargin=4pt,xrightmargin=4pt,
	aboveskip=0pt,belowskip=0pt,
	breaklines=true,breakindent=0pt,breakautoindent=false,
	columns=fullflexible,keepspaces=true,
	showstringspaces=false}
\lstdefinestyle{tracebox}{%
	basicstyle={\scriptsize\ttfamily},
	numbers=none,
	frame=single,framerule=0.5pt,framesep=0pt,
	rulecolor=\color{tracerule},
	backgroundcolor=\color{tracebg},
	framexleftmargin=4pt,framexrightmargin=4pt,
	framextopmargin=4pt,framexbottommargin=4pt,
	xleftmargin=4pt,xrightmargin=4pt,
	aboveskip=0pt,belowskip=0pt,
	breaklines=true,breakindent=0pt,breakautoindent=false,
	columns=fullflexible,keepspaces=true,
	showstringspaces=false}
\floatstyle{ruled}
\newfloat{listing}{tb}{lst}{}
\floatname{listing}{Listing}
\usepackage{booktabs}
\usepackage{xcolor}
\definecolor{gaincolor}{RGB}{0,132,80}
\newcommand{\gain}[1]{\,{\scriptsize\textcolor{gaincolor}{(+#1)}}}
\newcommand{\drop}[1]{\,{\scriptsize\textcolor{red}{(#1)}}}
\title{VTree: Question-Driven Growth of Hierarchical Memory for \\Agentic Long-Video Understanding}
\author{
    Xiaoyu Ma\textsuperscript{\rm 1},
    Lianyu Hu\textsuperscript{\rm 1},
    Zixuan Hu\textsuperscript{\rm 1},
    Zhizhen Wu\textsuperscript{\rm 1},
    Zeqin Liao\textsuperscript{\rm 1},
    Shengqian Qin\textsuperscript{\rm 2},
    Yang Liu\textsuperscript{\rm 1}
}
\affiliations{
    \textsuperscript{\rm 1}Nanyang Technological University, Singapore\\
    \textsuperscript{\rm 2}Shanghai Jiao Tong University, Shanghai, China\\
    ma0009yu@e.ntu.edu.sg, lianyu.hu@ntu.edu.sg, ZIXUAN014@e.ntu.edu.sg, c250082@e.ntu.edu.sg,\\
    zeqin.liao@ntu.edu.sg, qinshengqian@sjtu.edu.cn, yangliu@ntu.edu.sg
}

\begin{document}

\maketitle

\begin{abstract}
Long videos spanning tens of minutes to hours far exceed the context window of today's vision-language models, and the standard remedy of sparse uniform frame sampling discards exactly the fine-grained evidence that specific questions depend on. Existing video agents explore on demand but re-explore from scratch for every question and discard their exploration state once an answer is produced, so per-question cost scales with video length. Hierarchical-memory methods instead fix the scope of each depth in advance and deepen along a pre-committed pyramid rather than toward the question. We observe that a video's coarse structure (where its scenes and topics begin and end) is question-independent, whereas which part a question must zoom into, and how deep, is question-dependent. We propose VTree, which builds a coarse hierarchical memory tree over a video once and then, for each question, \emph{grows} branches only where that question needs depth. The growth is globally routed so any part of the timeline is reachable, sufficiency-gated so drilling stops when the evidence suffices, and query-conditioned so a branch unfolds into the sub-intervals that answer \emph{this} question. Crucially, every expansion is written back into the video memory, allowing later questions to reuse previously discovered structure and evidence without repeating the corresponding expansion process. Across five long-video understanding benchmarks, VTree consistently improves the accuracy of diverse vision-language backbones while substantially reducing the context length and execution time required for each question.
\end{abstract}

\section{Introduction}

\begin{figure}[!t]
\centering
\includegraphics[width=\columnwidth]{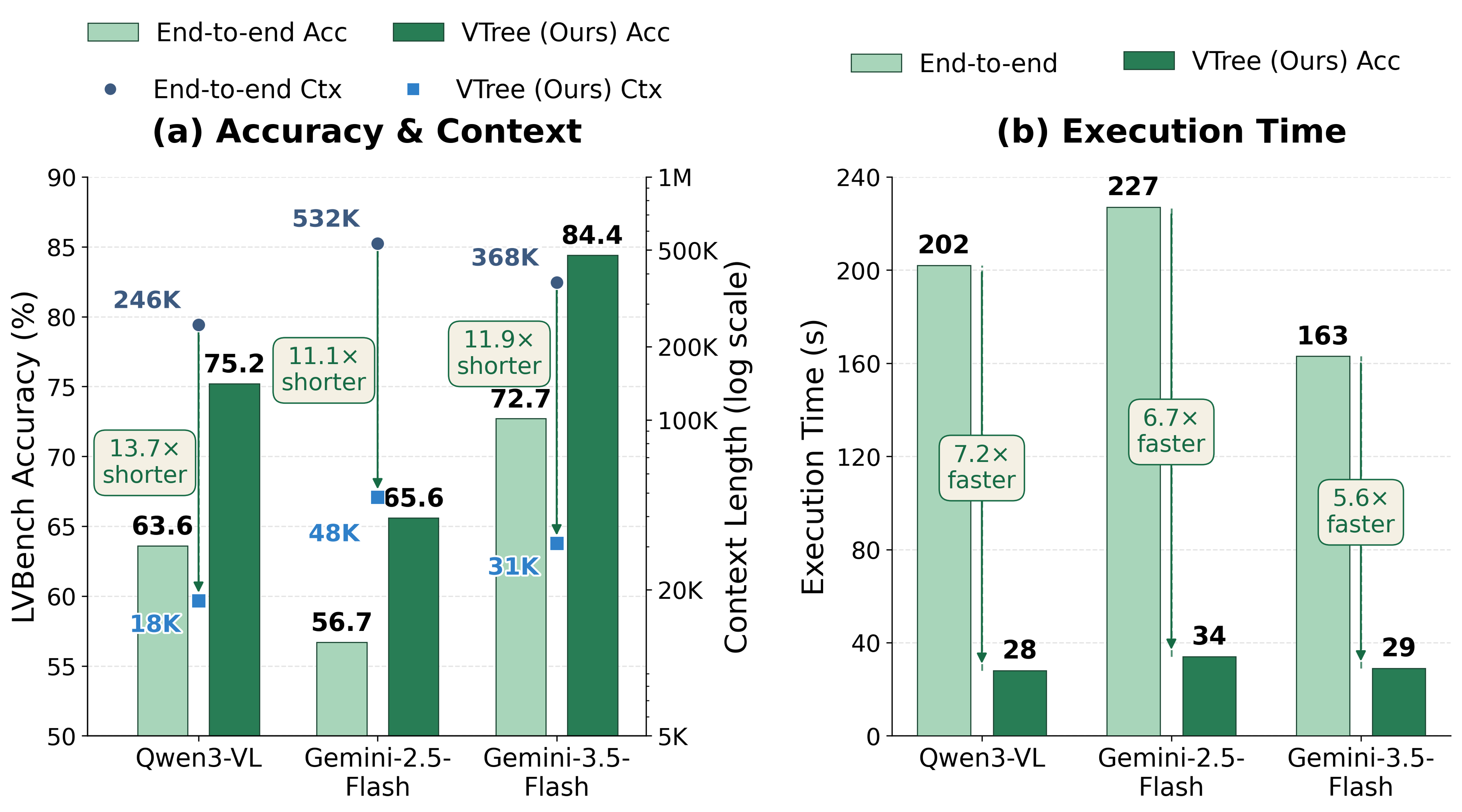}
\vspace{-0.7cm}
\caption{VTree raises accuracy while sharply cutting per-question cost. Across three backbones (Qwen3-VL, Gemini-2.5-Flash, Gemini-3.5-Flash) on LVBench: \textbf{(a)~Accuracy \& context}---VTree (green) outperforms feeding the video end-to-end (purple) while reading $11$--$14\times$ shorter context per question; \textbf{(b)~Execution time}---VTree answers $5.6$--$7.2\times$ faster in wall-clock time. Rather than ingesting the whole timeline, VTree grows its memory tree toward each question and reads only the branches the answer needs.}
\label{fig:teaser}
\vspace{-0.7cm}
\end{figure}

Understanding long videos with vision-language models (VLMs)~\cite{videochatgpt2023,videochat2023,videollava2023} is fundamentally a problem of scale. A video running for tens of minutes to several hours produces far more frames than a single model call can attend to, so passing the whole video as context, while increasingly feasible with very long-context models, remains computationally wasteful---the cost scales with video length and repeats for every question. The standard remedy, sparse uniform frame sampling, fits the budget but is blind to where an answer lies: a question about a brief action or a single spoken sentence depends on a short window that uniform sampling is likely to skip. Long-video question answering therefore hinges on reading the right parts of a video at the right granularity, not on reading more of it.

A promising line of work replaces fixed sampling with agents that use large language models (LLMs) to explore a video on demand---searching, captioning, or backtracking until they gather enough evidence to answer~\cite{dvd2025,videoagentwang2024,llovi2023}. This adaptivity helps, but it is paid for once per question: the clips an agent inspects and the intermediate state it builds are discarded as soon as the answer is produced, so the next question starts over and re-examines the same regions. A second line of work imposes a hierarchical memory over the timeline~\cite{videolucy2025,videotree2024,malmm2024}, but fixes the temporal scope of each depth in advance, deepening along a pre-committed pyramid rather than toward the specific question. Both families share the same conflation: they entangle what is question-independent with what is question-dependent.

We trace this to a single observation: where a video's scenes and topics begin and end---its coarse skeleton---is a property of the video and does not change from one question to the next, whereas which short window answers a particular question is a property of the question. Existing methods conflate the two, either re-exploring the video for every question or pre-committing the depth structure before the question is known. Separating them points to a different design: \textbf{build the skeleton once, and let each question \emph{grow the branches it needs} into that skeleton}---unfolding fine detail only where, and only as deep as, the question demands, and reaching any branch regardless of how it ranks. Crucially, the tree persists: every branch grown for one question remains available to subsequent questions about the same video.

We realize this idea in VTree. A hierarchical memory tree is built over a video once, producing coarse event nodes that form a trunk spanning the full timeline. For each question, VTree \emph{grows} the tree toward the answer through three cooperating levers: a router selects which branch to unfold and can reach \emph{any} node, not only the highest-scored ones, so evidence never hides below a cutoff; a sufficiency check stops drilling as soon as the collected evidence answers the question; and fine segmentation is \emph{query-conditioned}, so a branch unfolds into the sub-intervals that answer \emph{this} question, not a fixed pyramid. An LLM scores nodes to select what to read and what to expand, and read-pattern routing handles whole-timeline questions (Section~\ref{sec:method}). Above all, every expansion is \textbf{written back} into the tree, so the memory accumulates across the question stream: a branch grown for question $i$ becomes structure that questions $i{+}1,\dots,n$ reuse for free.

Our contributions are as follows. (i)~Question-conditioned branch expansion over a hierarchical video-memory tree: VTree grows the specific subtrees a question needs, globally routed so any part of the timeline is reachable and sufficiency-gated so drilling stops when the evidence is enough. (ii)Persistent video memory across a sequence of questions: query-driven expansions are written back into the tree, enabling subsequent questions to reuse previously discovered structure and evidence. (iii)~Supporting mechanisms that make this practical: an LLM that scores nodes to select what to read and what to expand, and read-pattern routing for timeline-global tasks. Across five long-video understanding benchmarks, VTree consistently improves diverse open- and closed-source backbones while substantially reducing the context processed and execution time per question(Figure~\ref{fig:teaser}).

\section{Related Work}

\subsection{Long Video Understanding}

Long-video understanding requires models to preserve broad temporal coverage while retaining sufficient visual detail for localized reasoning. Recent work has increasingly moved beyond simply enlarging the context window toward two complementary directions: constructing compact long-horizon representations and acquiring question-relevant evidence adaptively.

\textbf{Efficient long-horizon representations.}
Recent methods summarize or propagate video information under bounded memory. ReWind~\cite{rewind2025} maintains an instruction-conditioned memory through recurrent read, perceive, and write operations, and uses the resulting memory to retrieve a small number of high-resolution frames. SEAL~\cite{seal2025} represents long videos using compact scene, object, and action entities selected according to semantic relevance and diversity. For streaming settings, VideoLLaMB~\cite{videollamb2025} introduces recurrent memory bridges that propagate historical visual information across successive video segments. These approaches improve temporal coverage at manageable cost, but their retained representations are largely determined during a single sequential pass and may suppress visual details that are not prioritized at encoding time.

\textbf{Query-adaptive evidence acquisition.}
A second direction selects or retrieves visual evidence according to the input question. Flexible Frame Selection~\cite{ffs2025} learns a lightweight policy for selecting informative frames, while Adaptive Keyframe Sampling~\cite{aks2025} dynamically adjusts temporal sampling according to video content and query relevance. More recent work formulates frame selection as reinforcement learning~\cite{rlframeselect2026}, allowing the selector to optimize directly for downstream long-video reasoning. Retrieval-based systems such as DrVideo~\cite{drvideo2024} instead construct a coarse textual representation of the video, retrieve relevant intervals, and supplement them with detailed visual observations. Although these methods reduce redundant perception, they typically produce a question-specific subset of frames or retrieve independently indexed units without progressively refining a shared representation across multiple questions.

VTree combines global temporal coverage with adaptive local inspection. It constructs a reusable coarse representation once and writes query-driven refinements back into the same video memory, allowing previously acquired information to support subsequent questions.

\subsection{Agentic Exploration and Hierarchical Video Memory}

Beyond learned selection, recent systems cast long-video understanding as interactive evidence gathering, where an LLM or multimodal agent decides what to inspect. These methods divide into per-query agentic exploration and structured video-memory approaches.

\textbf{Per-query agentic exploration.}
VideoAgent~\cite{videoagentwang2024} established an iterative framework in which an LLM searches for relevant frames and accumulates evidence until it can answer the question. Deep Video Discovery (DVD)~\cite{dvd2025} provides an agent with search and inspection tools over a multi-granular video database. VCA~\cite{vca2025} uses curiosity-guided tree search to explore informative video segments, whereas LVAgent~\cite{lvagent2025} coordinates multiple MLLM agents over several rounds of perception and reasoning. VideoSeek~\cite{videoseek2026} further introduces tool-guided seeking operations for navigating long temporal contexts, while Ego-R1~\cite{egor1_2025} employs chain-of-tool-thought for reasoning over long egocentric videos. These methods enable adaptive and self-directed evidence acquisition, but the resulting exploration trajectory is generally scoped to the current question rather than retained as an evolving representation of the video.

\begin{figure*}[t]
\centering
\includegraphics[width=\textwidth]{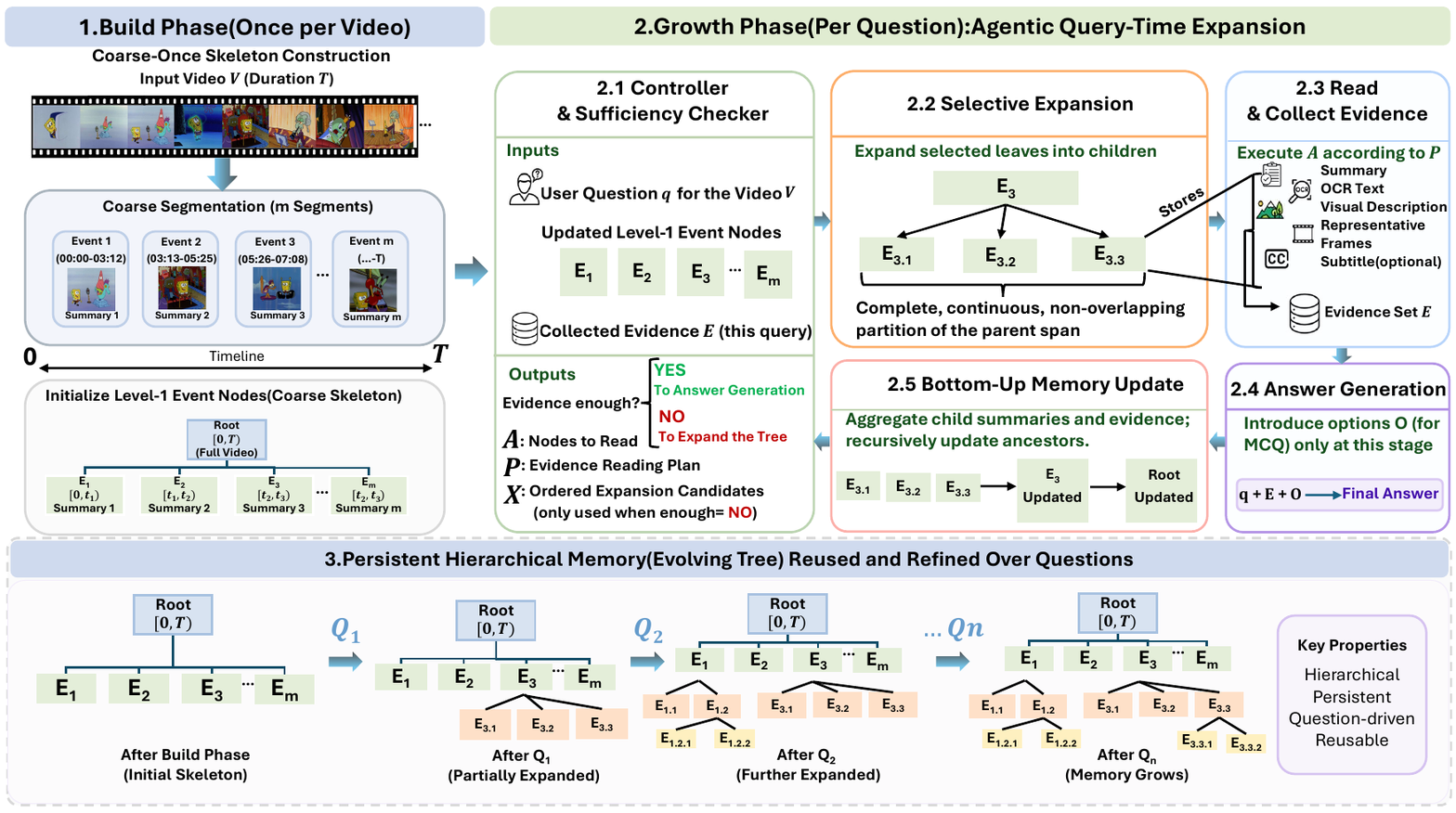}
\vspace{-2cm}
\caption{
Overview of VTree.
\textbf{(1) Build phase:} a question-independent level-$1$ event
skeleton is constructed once per video.
\textbf{(2) Growth phase:} for each question, the controller routes over
the updated level-$1$ nodes, reuses stored descendants, and expands an
unexplored leaf only when finer evidence is needed. Newly acquired
information is propagated bottom-up.
\textbf{(3) Persistent memory:} the refined hierarchy is retained across
questions for subsequent reuse.
}
\label{fig:framework}
\vspace{-0.4cm}
\end{figure*}

\textbf{Structured and hierarchical video memory.}
A complementary line of work explicitly organizes video information for coarse-to-fine access. VideoTree~\cite{videotree2024} constructs a query-adaptive hierarchy through visual clustering and relevance-guided expansion. VideoLucy~\cite{videolucy2025} performs memory backtracking over predefined temporal scales, while VideoARM~\cite{videoarm2025} interleaves agentic reasoning with the dynamic construction of hierarchical multimodal memory. MemDreamer~\cite{memdreamer2026} similarly separates perception and reasoning through a hierarchical graph representation. These methods introduce useful temporal structure, but their hierarchies are typically predefined or constructed within the processing of an individual query.

VTree maintains a video-centric hierarchy across a sequence of questions. It builds a coarse skeleton once, expands only the branches required by each query, and preserves the resulting nodes and evidence for future reuse.

\section{Method}
\label{sec:method}

VTree maintains a persistent hierarchical memory for each video and incrementally refines it as questions arrive. It operates in two phases (Figure~\ref{fig:framework}). A \emph{build} phase runs once per video and constructs a coarse, question-independent temporal skeleton through a whole-video VLM pass. A \emph{growth} phase runs for each question: an LLM controller examines the current tree, determines whether the available information is sufficient, and expands selected nodes when finer evidence is needed. Every expansion is written back into the shared tree, allowing later questions to access previously acquired structure and evidence when relevant.

The tree has no predefined final form: a question may be answered directly from coarse summaries, may read evidence from existing nodes, or may trigger several rounds of finer expansion. The memory thus evolves with the question sequence.

\subsection{Tree Representation}

Let a video $V$ have duration $T$. VTree maintains a hierarchical memory
\begin{equation}
\mathcal{T}=(\mathcal{N},\mathcal{S},\mathcal{E}),
\label{eq:tree}
\end{equation}
where $\mathcal{N}$ denotes tree nodes, $\mathcal{S}$ temporal segments, and $\mathcal{E}$ evidence items.

Each node $n\in\mathcal{N}$ stores its parent and children, depth
$\ell(n)$, temporal span $[s_n,e_n)$, node type
$t(n)\in\{\textsc{root},\textsc{event},\textsc{subevent}\}$,
a summary, an expansion status, and pointers to its associated segment
and evidence. Each segment stores its temporal span, summary, transcript,
OCR text, visual description, and representative frames. An evidence
item is any of these segment-level contents retrieved at query time.

The root at depth $0$ spans the full video. Level-$1$ \textsc{event} nodes form the coarse skeleton, while deeper \textsc{subevent} nodes are created during query-time growth. The maximum tree depth is $D$, set to $4$ in our experiments.

The current leaf set is
\begin{equation}
\mathcal{L}(\mathcal{T})
=
\{n\in\mathcal{N}
\mid
\mathrm{children}(n)=\emptyset,\,
n\neq\textsc{root}\}.
\label{eq:leaves}
\end{equation}
A leaf is a node that has not yet been expanded. Since only query-relevant branches are expanded, leaves can appear at different depths of the tree.

For every expanded node $n$ with ordered children
$\{c_1,\ldots,c_K\}$, the children form a complete, non-overlapping partition of the parent:
\begin{equation}
s_{c_1}=s_n,\qquad
e_{c_K}=e_n,\qquad
e_{c_i}=s_{c_{i+1}}
\quad \forall i<K.
\label{eq:partition}
\end{equation}
Thus, query-conditioned expansion changes the temporal granularity but never removes any part of the parent interval from future access.

\subsection{Coarse-Once Skeleton Construction}

The video is decoded once at one frame per second, and the sampled
frames are cached for subsequent processing. The build phase then
performs a single question-independent VLM pass to obtain a coarse
overview of the video. This cost is incurred once per video rather than
repeated for every question.

Given a video of duration $T$, the VLM identifies semantic event
boundaries and partitions the timeline into coarse topic-level segments.
The number and duration of these segments are determined by the video
content rather than fixed in advance. Compared with uniform temporal
partitioning, this content-adaptive segmentation is less likely to split
a coherent event or merge semantically unrelated content. Each segment
is assigned a temporal interval and a concise summary. The intervals are
normalized to form a complete, non-overlapping partition of $[0,T)$.

Each coarse segment becomes a level-$1$ \textsc{event} node that stores
its summary and a pointer to the corresponding video clip. Dense frame
descriptions, transcripts, OCR results, and other fine-grained evidence
are deferred until a question requires refinement of that branch. A
global video summary is constructed by temporally concatenating the
level-$1$ summaries under a fixed length limit.

The resulting shallow skeleton captures the coarse semantic organization
of the video. Branches that are never selected by any question remain at
this summary-level representation, avoiding unnecessary fine-grained
processing.

\subsection{Global Routing and Sufficiency Checking}

For each question $q$, the controller first performs global routing over
the level-$1$ event nodes
$\mathcal{N}^{(1)} = \mathrm{children}(\mathrm{root})$,
which form a complete, non-overlapping partition of the full video
timeline. The controller receives the question, the temporal span and
summary of each level-$1$ event node, and the evidence collected for the
current question. Because these summaries are updated after previous
query-time expansions, they provide a compact view of both the original
coarse structure and the information accumulated within each branch.

Answer options are not provided during routing, branch refinement, or
sufficiency checking; they are introduced only during final answer
generation. This prevents temporal localization from being directly
biased by option wording.

All level-$1$ nodes are presented in temporal order without an
embedding-based top-$k$ pre-filter or a fixed node-count cutoff. Global
routing therefore retains complete coarse coverage of the video while
avoiding the cost and redundancy of presenting the entire evolving tree
to the controller.

The controller returns
$(\delta,\mathcal{A},\mathcal{X},\mathcal{P})$,
where $\delta\in\{0,1\}$ is the sufficiency decision, with $\delta=1$
indicating that the collected information is sufficient.
$\mathcal{A}\subseteq\mathcal{N}^{(1)}$ contains level-$1$ branches
whose stored evidence should be read,
$\mathcal{X}\subseteq\mathcal{N}^{(1)}$ is an ordered set of branches
selected for further refinement, and $\mathcal{P}$ specifies which
summaries, textual evidence, or frames should be retrieved.

If the current evidence is insufficient, VTree performs local
coarse-to-fine refinement only within the selected level-$1$ branches.
Stored descendants created by earlier questions are traversed and reused
before any new video perception is invoked. A new query-conditioned
expansion is performed only when refinement reaches an unexpanded leaf
whose current information remains insufficient.

\subsection{Query-Conditioned Branch Refinement}

When the evidence collected through level-$1$ global routing is
insufficient, the controller selects a level-$1$ branch
$b\in\mathcal{X}$ for local refinement. Starting from $b$, VTree
traverses its stored descendants in a coarse-to-fine manner, using their
summaries and available evidence to identify the most relevant
subbranches. Descendants created by previous questions are directly
reused during this process without repeating video perception.

If the stored information remains insufficient and refinement reaches
an unexpanded leaf $x$ with $\ell(x)<D$, VTree performs a new
query-conditioned expansion. The expansion receives the question, the
temporal clip associated with $x$, and its stored textual information,
and analyzes the clip at a higher visual density than that used during
the coarse build phase.

The VLM partitions the span of $x$ into finer subevents. Their
boundaries and summaries are conditioned on the current question so
that additional resolution is allocated around potentially relevant
temporal changes. Nevertheless, the resulting intervals must form a
complete, continuous, and non-overlapping partition of the parent span.
For a node $x$ spanning $[s_x,e_x)$, expansion produces
\begin{equation}
\{[s_1,e_1),\ldots,[s_K,e_K)\},
\label{eq:expansion_output}
\end{equation}
subject to Equation~\ref{eq:partition}. We allow at most twenty children
per expansion, with a minimum duration of three seconds whenever the
duration of the parent node permits.

Each interval becomes a \textsc{subevent} node at depth
$\ell(x)+1$. It initially stores a segment summary, the available
textual evidence, and three representative frames. The newly created
children are written back into the persistent tree and become
immediately available to both the current question and subsequent
questions. Because they completely partition the parent span, future
questions may continue refining any portion of the branch, including
regions that were not relevant to the question that originally
triggered the expansion.

\subsection{Bottom-Up Memory Update}

After a new expansion, the summaries and evidence of the resulting
children are aggregated into an enriched representation of their
parent. This update is recursively propagated through the selected
branch to the corresponding level-$1$ event node, under a bounded
summary length. Detailed information remains stored in the descendant
nodes, while each updated ancestor provides a compact representation of
the information currently available within its subtree.

Every new question begins with global routing over the level-$1$ event
nodes. Therefore, propagating newly acquired information to the
corresponding level-$1$ summary directly affects subsequent branch
selection. Once an updated level-$1$ branch is selected, VTree can reuse
its stored descendants and progressively refine the branch without
repeating previously completed video perception.

\subsection{Sufficiency-Gated Query Processing}

For each question, VTree alternates between routing, evidence collection, and expansion. The maximum depth $D$ limits the finest branch granularity, while the drill budget $R$ limits the expansions per question; we use $R=5$.

Each round, the controller reads the current leaves and collected evidence. If sufficient, it returns a reading plan drawing on summaries, textual evidence, or one to several frames across intervals; otherwise the highest-priority expandable node is grown and the question is re-evaluated.

\begin{algorithm}[t]
\caption{VTree sufficiency-gated query-time refinement}
\label{alg:query}
\begin{algorithmic}[1]
\REQUIRE tree $\mathcal{T}$; question $q$; options $O$;
expansion budget $R$; depth limit $D$
\ENSURE answer $a$

\STATE $E\gets\emptyset$
\STATE $\mathcal{N}^{(1)}
\gets\textsc{Children}(\textsc{Root}(\mathcal{T}))$

\FOR{$r=0$ to $R$}

    \algstage{Global routing and sufficiency checking}

    \STATE $(\delta,\mathcal{A},\mathcal{X},\mathcal{P})
    \gets\textsc{GlobalControl}
    (q,\mathcal{N}^{(1)},E)$

    \algstage{Evidence collection}

    \STATE $E\gets\textsc{Merge}
    \bigl(E,\textsc{Collect}(\mathcal{A},\mathcal{P})\bigr)$

    \IF{$\delta=1$ \OR $r=R$}
        \STATE \textbf{break}
    \ENDIF

    \algstage{Local branch refinement}

    \STATE $x\gets\textsc{FirstBranch}(\mathcal{X})$
    \IF{$x=\emptyset$}
        \STATE \textbf{break}
    \ENDIF

    \STATE $(E,\mathcal{T})
    \gets\textsc{RefineBranch}
    (x,q,E,\mathcal{T},D)$
\ENDFOR

\algstage{Answer generation and write-back}

\STATE $a\gets\textsc{Answer}(q,O,E)$
\STATE $\textsc{Persist}(\mathcal{T})$
\RETURN $a$
\end{algorithmic}
\vspace{-0.05cm}
\end{algorithm}

\subsection{Evidence Collection and Persistent Write-Back}

Following the controller's reading plan, the evidence collector uses node summaries directly or retrieves denser textual and visual evidence from selected clips. Items are deduplicated by type and half-second-bucketed span, sorted by time, and formatted as an ordered list of timestamps, text, and frame paths. The answerer receives the question and this evidence; multiple-choice options are supplied only here, and a constrained single-token output predicts the option.

After each question, the tree is serialized to a video-specific directory, and subsequent questions about the same video load the full accumulated state (nodes, hierarchical relations, summaries, evidence, and ancestor updates). Previous answers, options, and raw question text are not stored, so the memory is video-centric rather than conversational: later questions reuse discovered structure and evidence without earlier answers as context, and a new video initializes a separate tree. This persistence amortizes prior segmentation, perception, and expansion costs whenever later questions need overlapping evidence, though the saving is not monotonic since routing over the accumulated tree also incurs cost.

\section{Experiments}

\begin{table*}[t]
\centering
\small
\setlength{\tabcolsep}{3pt}
\begin{tabular}{l lll llll ll}
\toprule
\multicolumn{1}{l}{Methods} & EgoSchema & LVBench & MLVU & \multicolumn{4}{c}{Video-MME} & \multicolumn{2}{c}{LongVideoBench} \\
\cmidrule(lr){5-8}\cmidrule(lr){9-10}
 & Val & Overall & Test & Short & Medium & Long & Overall & Val & Long \\
\midrule
Seed-1.5-VL~\cite{seed15vl2025}          & --   & 64.6 & 82.1 & -- & -- & --   & 77.9 & 74.4 & --   \\

GPT-4.1~\cite{gpt41_2025} & 72.0 & 63.4 & --   & -- & -- & --   & 72.0 & --   & --   \\
GPT-5~\cite{gpt5card2025} &-- & 66.5 & --   & -- & -- & 67.9   & 72.0 & --   & 64.5   \\
\midrule
VideoTree~\cite{videotree2024}             & 66.2 & 28.8 & 60.4 & -- & -- & 54.2 & 60.6   & --   & --   \\
SiLVR~\cite{silvr2025}                & 44.4 & --   & 45.2 & -- & -- & --   & 74.1   & --   & --   \\
VideoLucy~\cite{videolucy2025}            & --   & 58.8 & 76.1 & -- & -- & 66.8 & 72.5   & --   & --   \\
LVAgent~\cite{lvagent2025}              & 82.9   & 80.0   & 83.9 &88.9 &82.0 & 74.3 & 81.7   & 80.0 & --   \\
DVD~\cite{dvd2025}& -- & 74.2 & --   & -- & -- & 67.3 & --     & 71.6 & 68.6 \\
AVP~\cite{avp2025}& 73.6 & 74.8 & 84.3 & -- & -- & 81.9 & 85.3 & 73.4 & 70.0 \\
VideoSeek~\cite{videoseek2026} & -- & 68.4 & -- & -- & -- & 70.1 & 85.3 & -- & 73.5 \\
VideoARM~\cite{videoarm2025} & 76.2 & 68.4 & -- &85.8 & 81.3& 81.2 & 82.8 & 78.0 & 76.4 \\
\midrule
Qwen3-VL & 75.9 & 63.6 & 73.5 & 79.9 &78.4 & 71.2 & 76.5 & 71.4 & 67.1 \\
VTree + Qwen3-VL   & \textbf{84.5}\gain{8.6} & \textbf{75.2}\gain{11.6} & \textbf{79.4}\gain{5.9} & \textbf{82.6}\gain{2.7} & \textbf{80.6}\gain{2.2} & \textbf{78.3}\gain{7.1} & \textbf{80.5}\gain{4.0} & \textbf{75.1}\gain{3.7} & \textbf{70.8}\gain{3.7} \\
\cmidrule(lr){1-10}
GPT-4o               & 70.4 & 48.9 & 54.9 & 80.0&70.3 & 65.3 & 71.9 & 66.7 & 60.9 \\
VTree + GPT-4o   & \textbf{79.5}\gain{9.1} & \textbf{64.2}\gain{15.3} & \textbf{62.4}\gain{7.5} & \textbf{80.6}\gain{0.6} & \textbf{74.6}\gain{4.3} & \textbf{71.3}\gain{6.0} & \textbf{75.5}\gain{3.6} & \textbf{70.1}\gain{3.4} & \textbf{65.2}\gain{4.3} \\
\cmidrule(lr){1-10}
Gemini-2.5-Flash & 72.5 & 56.7 & 72.4 &81.2 &72.3 & 69.1 & 74.2 & 66.2 & 61.8 \\
VTree + Gemini-2.5-Flash & \textbf{81.2}\gain{8.7} & \textbf{65.6}\gain{8.9} & \textbf{74.5}\gain{2.1} & \textbf{84.3}\gain{3.1} & \textbf{83.5}\gain{11.2} & \textbf{78.2}\gain{9.1} & \textbf{82.0}\gain{7.8} & \textbf{72.8}\gain{6.6} & \textbf{66.2}\gain{4.4} \\
\cmidrule(lr){1-10}
Gemini-2.5-Pro& 72.8 & 72.0 & 79.6 & 86.7&82.9 & 75.9 & 82.4 & 71.0 & 68.6 \\
VTree + Gemini-2.5-Pro & \textbf{86.4}\gain{13.6} & \textbf{82.5}\gain{10.5} & \textbf{86.1}\gain{6.5} & \textbf{89.4}\gain{2.7} & \textbf{87.5}\gain{4.6} & \textbf{84.7}\gain{8.8} & \textbf{87.2}\gain{4.8} & \textbf{78.5}\gain{7.5} & \textbf{76.3}\gain{7.7} \\
\cmidrule(lr){1-10}
Gemini-3.5-Flash & 79.4 & 72.7 & 88.2 &\textbf{93.3} &90.1 & 80.3 & 87.9 & 71.4 & 69.2 \\
VTree + Gemini-3.5-Flash & \textbf{88.5}\gain{9.1} & \textbf{84.4}\gain{11.7} & \textbf{89.5}\gain{1.3} & 93.2\drop{-0.1} & \textbf{90.6}\gain{0.5} & \textbf{86.5}\gain{6.2} & \textbf{90.1}\gain{2.2} & \textbf{81.6}\gain{10.2} & \textbf{78.4}\gain{9.2} \\

\bottomrule
\end{tabular}
\caption{Main results across long-video QA benchmarks (accuracy, \%). Within each backbone/VTree pair, the higher value per column is in \textbf{bold}; colored \textcolor{gaincolor}{(+x.x)} marks VTree's gain over its backbone.}
\label{tab:main}
\vspace{-0.2cm}
\end{table*}

\subsection{Experimental Setup}

\noindent\textbf{Evaluation Benchmarks.}
We evaluate on five long-video benchmarks. LVBench~\cite{lvbench2025}
focuses on extremely long videos, and we report the number of evaluated
clips out of 103. Video-MME~\cite{videomme2025} covers short-, medium-,
and long-duration videos, with results reported by duration, domain. MLVU~\cite{mlvu2025} contains nine long-video tasks, reported
individually. We evaluate LongVideoBench~\cite{longvideobench2024} on
its validation set and EgoSchema~\cite{egoschema2023} on its labelled
500-question subset using overall accuracy.

\noindent\textbf{Implementation Details.}
In the build phase, a single whole-video pass segments the timeline into
coarse events to initialize the persistent skeleton. During query-time
growth, selected branches are progressively refined with a maximum
budget of five expansion rounds. We set the maximum tree depth to
$D=4$ as a safety constraint; no evaluated video reaches this limit.
Node selection and evidence sufficiency are determined by the LLM
controller. We provide no subtitle files and follow the
without-subtitles protocol throughout, allowing the backbone to
transcribe embedded audio as ASR evidence. Within each run, the same
backbone performs evidence extraction, sufficiency checking, and answer
generation, and we report its exact version string. We instantiate
VTree with the closed-source GPT-4o, Gemini-2.5-Flash,
Gemini-2.5-Pro, and Gemini-3.5-Flash families, as well as the
open-source Qwen3-VL. The full prompt set, together with the
direct-backbone baseline prompt, is reproduced in
Appendix~\ref{sec:appendix_prompts}.

\subsection{Main Results}
Table~\ref{tab:main} reports accuracy across the five benchmarks. Attaching VTree to a backbone lifts that backbone on nearly every column, and the strongest configuration, VTree with Gemini-3.5-Flash, sets the leading score on each benchmark. On LVBench it reaches $84.4$, an $11.7$-point gain over the same Gemini-3.5-Flash backbone and $4.4$ points above the best specialized long-video agent (LVAgent, $80.0$). On MLVU and the overall Video-MME split it attains $89.5$ and $90.1$, and on the more demanding Video-MME long split it reaches $86.5$, a $6.2$-point improvement over its backbone; on LongVideoBench it leads at $81.6$ and $78.4$ on the validation and long subsets. The largest per-benchmark lifts surface where the backbone is weakest: VTree raises GPT-4o by $15.3$ points on LVBench and Gemini-2.5-Pro by $13.6$ points on EgoSchema. Even on EgoSchema, whose egocentric viewpoint stresses temporal reasoning, the best configuration secures $88.5$. These gains hold across both open-source (Qwen3-VL) and closed-source (GPT-4o and the Gemini families) engines, indicating that the improvement stems from the persistent tree rather than any single backbone. 

\subsection{Ablations}

Unless otherwise specified, all ablations use Gemini-2.5-Flash on
LVBench.

\subsubsection{Coarse Segmentation Strategy}

We compare random intervals (60--180\,s), fixed 120\,s windows,
DDM-Net~\cite{tang2022progressive}, and our LLM-based segmentation procedure
(\S\ref{sec:method}).

\begin{table}[t]
\centering
\small
\begin{tabular}{lc}
\toprule
Segmentation Strategy & Accuracy (\%) \\
\midrule
Random intervals (60--180\,s) & 42.9 \\
Fixed-time windows (120\,s)   & 58.7 \\
DDM-Net~\cite{tang2022progressive} & 61.5 \\
LLM-based (ours)              & \textbf{65.6} \\
\bottomrule
\end{tabular}
\caption{Coarse segmentation ablation on LVBench.}
\label{tab:abl_seg}
\vspace{-0.1cm}
\end{table}

As shown in Table~\ref{tab:abl_seg}, semantic segmentation substantially
outperforms blind temporal partitioning. Our LLM-based strategy improves
over random intervals, fixed windows, and DDM-Net by 22.7, 6.9, and
4.1 points, respectively. Since subsequent refinement operates within
the level-$1$ branches, errors introduced during coarse segmentation
cannot be recovered by deeper expansion.

\begin{figure*}[t]
\centering
\includegraphics[width=\textwidth]{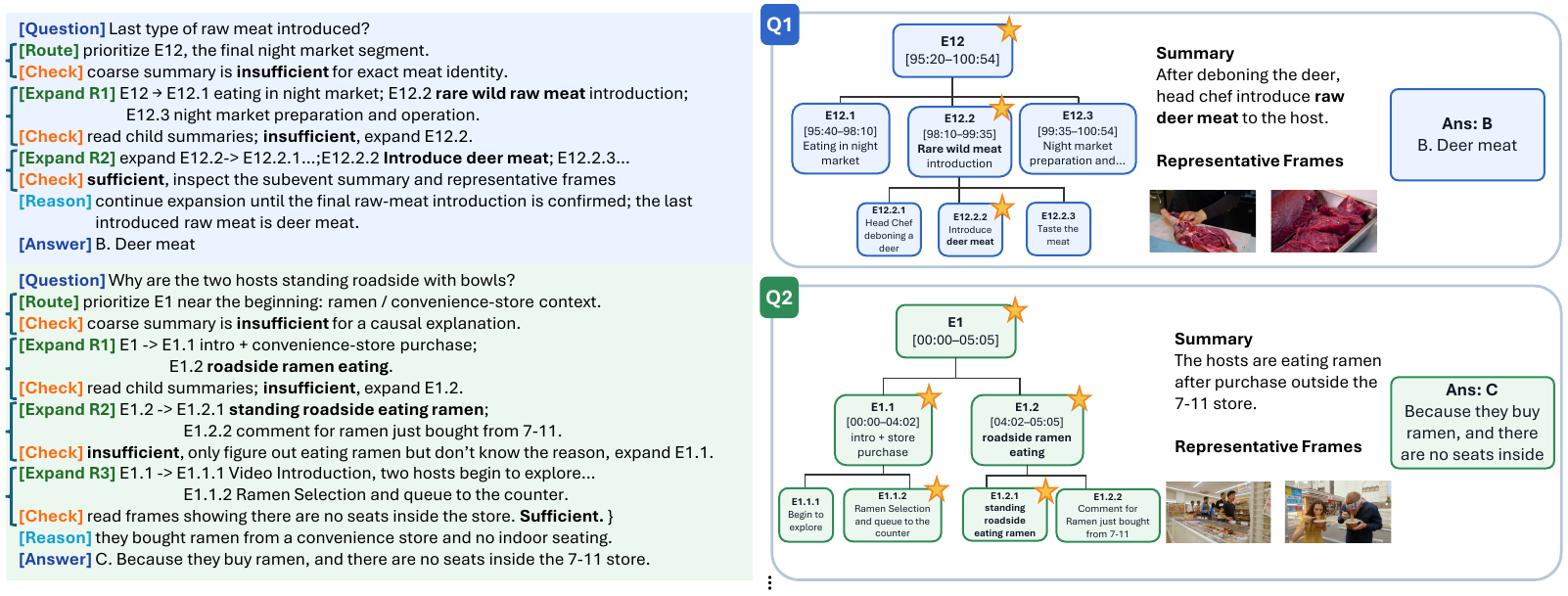}
\vspace{-0.5cm}
\caption{Qualitative examples of the Route--Check--Expand process. VTree
routes to a relevant branch, checks evidence sufficiency, and expands
the branch only when finer detail is required.}
\label{fig:visualization}
\vspace{-0.35cm}
\end{figure*}

\subsubsection{Persistent Write-Back}

We compare resetting all query-conditioned expansions after each
question with retaining expansions, evidence, and ancestor updates across
questions. All other configurations are identical.

\begin{table}[t]
\centering
\small
\setlength{\tabcolsep}{3.5pt}
\begin{tabular}{@{}lcccc@{}}
\toprule
& \multicolumn{2}{c}{LVBench}
& \multicolumn{2}{c}{Video-MME Long} \\
\cmidrule(lr){2-3}
\cmidrule(lr){4-5}
Protocol
& Acc. (\%) & Time (s)
& Acc. (\%) & Time (s) \\
\midrule
Reset per question
& 62.4 & 41
& 76.3 & 53 \\
Persistent write-back
& \textbf{65.6} & \textbf{34}
& \textbf{78.2} & \textbf{47} \\
\bottomrule
\end{tabular}
\caption{Cross-question persistence ablation. Time is averaged per
question.}
\label{tab:abl_persistence}
\vspace{-0.3cm}
\end{table}

Persistent write-back improves accuracy by 3.2 points on LVBench and
1.9 points on Video-MME Long, while reducing average execution time
from 41 to 34 seconds and from 53 to 47 seconds, respectively. Freezing
parent summaries while retaining the expanded tree reduces LVBench
accuracy from 65.6\% to 63.5\%, further confirming the benefit of
bottom-up updates for subsequent level-$1$ routing.

\subsubsection{Query-Time Expansion Budget}

We vary the maximum number of query-time expansion rounds from zero
(coarse-only) to unlimited.

\begin{table}[t]
\centering
\small
\begin{tabular}{lc}
\toprule
Max Expansion Rounds & Accuracy (\%) \\
\midrule
0 (coarse only) & 37.5 \\
1 round         & 54.2 \\
3 rounds        & 61.5 \\
5 rounds        & \textbf{65.6} \\
Unlimited       & 65.8 \\
\bottomrule
\end{tabular}
\caption{Query-time expansion budget ablation on LVBench.}
\label{tab:abl_rounds}
\end{table}

Table~\ref{tab:abl_rounds} shows diminishing returns as the expansion
budget increases. One round improves the coarse-only result by 16.7
points, while three rounds add another 7.3 points. Increasing the budget
from five rounds to unlimited yields only 0.2 additional points,
indicating that five rounds provide a favorable accuracy--efficiency
trade-off. The sufficiency gate allows simple questions to stop early
while allocating additional refinement rounds to harder cases.

\subsubsection{Question-Order Sensitivity}

Because expansions persist across questions, the memory may depend on
question order. We randomly select 20 LVBench videos containing 300
questions, shuffle the question order, and repeat the evaluation five
times under the same persistent configuration.

\begin{table}[t]
\centering
\small
\setlength{\tabcolsep}{1.5pt}
\begin{tabular*}{\columnwidth}{
@{\extracolsep{\fill}}lccccccc@{}}
\toprule
& Backbone
& \multicolumn{5}{c}{VTree Runs}
& Mean$\pm$Std. \\
\cmidrule(lr){3-7}
& Baseline & 1 & 2 & 3 & 4 & 5 & \\
\midrule
Acc. (\%)
& 59.3
& 67.3
& 66.7
& 69.3
& 68.7
& 68.3
& \textbf{68.1$\pm$1.1} \\
\bottomrule
\end{tabular*}
\caption{Question-order sensitivity on 20 LVBench videos with 300
questions.}
\label{tab:order_sensitivity}
\vspace{-0.2cm}
\end{table}

VTree obtains an average accuracy of 68.1\% with a standard deviation of
1.1 points. Results range from 66.7\% to 69.3\%, giving a maximum
difference of 2.6 points across question orders. This limited variation
suggests that performance is not strongly dependent on a particular
ordering, although the persistent tree is shaped by previously processed
questions. This subset experiment analyzes order sensitivity rather than
replacing the full LVBench evaluation.

\subsection{Analysis}

\noindent\textbf{Qualitative Analysis.}
Figure~\ref{fig:visualization} illustrates the Route--Check--Expand
workflow on two representative questions. VTree first routes to a
relevant level-$1$ branch, checks whether the available summaries and
evidence are sufficient, and progressively refines only the necessary
subevents. In the first example, successive expansion localizes the
fine-grained event identifying the final raw meat. In the second, VTree
combines evidence from multiple subevents to explain why the hosts are
eating by the roadside. These examples demonstrate how adaptive
refinement supports both precise temporal localization and multi-step
reasoning.

\noindent\textbf{Efficiency Analysis.}
Figure~\ref{fig:teaser} compares VTree with end-to-end inference on
LVBench across three backbones. By selectively refining branches instead
of processing the complete video, VTree reduces context usage by
$11$--$14\times$ and achieves a $5.6$--$7.2\times$ speedup. These
reductions are accompanied by consistent accuracy improvements over
full-video inference, showing that targeted evidence acquisition can
improve both computational efficiency and answer quality.

\section{Conclusion}

VTree separates question-independent structure from question-dependent
detail by constructing a coarse hierarchical memory tree once per video
and refining selected branches on demand. Query-conditioned expansions
are written back into the persistent tree, allowing subsequent questions
to reuse previously acquired structure and evidence. Through routing,
sufficiency-gated refinement, and bottom-up memory updates, VTree
improves long-video question answering across five diverse benchmarks
while processing substantially less context than direct end-to-end
inference. Future work will investigate long-term tree growth,
robustness over extended question streams, and adaptive memory
compression.

\bibliography{aaai2027}

\clearpage
\appendix
\raggedbottom
\section*{Supplementary Material}

\section{Prompts}
\label{sec:appendix_prompts}

This appendix reproduces every prompt that VTree issues to a model, so that
the pipeline of Section~\ref{sec:method} can be reproduced without access to
our code. Prompts are given verbatim, with \texttt{\{placeholder\}} fields
left in place; the value bound to each field is stated beneath its box. No
other component of VTree calls a model.

\subsection{Prompt Inventory and Call Budget}

VTree uses six prompts, listed in Table~\ref{tab:prompt_inventory} in
pipeline order. A seventh prompt, B1, is not part of VTree: it is the
direct-backbone baseline that produces the backbone rows of
Table~\ref{tab:main}, and is included here so that the comparison can be
checked.

\begin{table}[h]
\centering
\small
\setlength{\tabcolsep}{3pt}
\begin{tabular}{@{}llll@{}}
\toprule
ID & Prompt & Phase & Calls \\
\midrule
P1 & Coarse segmentation       & Build     & $1$ per video \\
P2 & Segment analysis          & On demand & $\le 1$ per segment \\
P3 & Controller \& sufficiency & Growth    & $\le R{+}1=6$ \\
P4 & In-branch descent         & Growth    & $\le R(D{-}1)=15$ \\
P5 & Branch expansion          & Growth    & $\le R=5$ \\
P6 & Answer generation         & Answer    & $1$ \\
\midrule
B1 & Direct-backbone baseline  & Baseline  & $1$ (not VTree) \\
\bottomrule
\end{tabular}
\caption{The prompt inventory. Build-phase calls are amortized over all
questions asked about a video; growth- and answer-phase counts are per
question, with $R=5$ the drill budget and $D=4$ the depth limit of
Algorithm~\ref{alg:query}.}
\label{tab:prompt_inventory}
\end{table}

Each round of Algorithm~\ref{alg:query} issues one P3 call. When that call
reports $\delta=0$, the selected branch is descended with one P4 call per
level traversed, and a single P5 call fires if the descent terminates at an
unexpanded leaf. Writing $b_q$ for the number of segments whose dense
evidence is materialized for the first time while answering $q$, the number
of model calls is bounded by
\begin{equation}
N(q)\;\le\;
\underbrace{(R{+}1)}_{\mathrm{P3}}
+\underbrace{R(D{-}1)}_{\mathrm{P4}}
+\underbrace{R}_{\mathrm{P5}}
+\underbrace{1}_{\mathrm{P6}}
+\underbrace{b_q}_{\mathrm{P2}},
\label{eq:budget}
\end{equation}
that is $27+b_q$ at $R=5$ and $D=4$. Equation~\ref{eq:budget} is loose in
both of its growth terms. The P4 term counts a full descent from level $1$
to the depth limit on every round, which can only occur once earlier
questions have already grown the branch that deep; on the first question
about a video every level-$1$ node is a leaf and the descent term is zero.
The P5 term assumes the gate never fires, whereas
Table~\ref{tab:abl_rounds} shows that lifting the drill budget from five
rounds to unlimited moves accuracy by $0.2$ points, so few questions reach
the cap. The build term is paid once: over a stream
$\mathcal{Q}=(q_1,\dots,q_n)$ on one video the total is
$1+\sum_{i}N(q_i)$, and the per-question average falls as $n$ grows
because $b_{q_i}$ and the P5 count both decay once a region has been
expanded.

\subsection{Build Phase}

\noindent\textbf{P1. Coarse segmentation.} Issued once per video, with the
whole video attached, to produce the level-$1$ event skeleton.

\begin{pbox}
\begin{lstlisting}[style=promptbox]
Identify {n_min}-{n_max} major scene or topic
segments in this video. Place each boundary where
the scene or the topic actually changes: the number
of segments and the duration of each segment must
follow the content, not a fixed schedule.
Reply ONLY with a JSON array:
[{"start":0.0,"end":30.0,"summary":"..."},...]
\end{lstlisting}
\end{pbox}

\noindent
\texttt{\{n\_min\}} scales with the duration $T$ of the video and
\texttt{\{n\_max\}}$=2\,\texttt{\{n\_min\}}$, so that the hint tracks how
long the video is instead of imposing a constant number of segments. The
range is a hint on the granularity, not a quota: the model returns however
many boundaries the content carries, and the returned count is what becomes
the level-$1$ width $m$. Because the reply is free to overlap or to leave
gaps, the parsed intervals $\{[\tilde s_i,\tilde e_i)\}_{i=1}^{m}$ are
sorted by $\tilde s_i$ and snapped before they become nodes,
\begin{equation}
s_1\!=\!0,\quad
e_m\!=\!T,\quad
e_i\!=\!s_{i+1}\!=\!\tilde s_{i+1}\ \ \forall i<m,
\label{eq:normalize}
\end{equation}
which makes the level-$1$ nodes satisfy Equation~\ref{eq:partition} with the
root as their parent. Each summary is carried over unchanged onto the node
whose interval it described.

\noindent\textbf{P2. Segment analysis.} Issued when a question first
requires dense evidence from a segment, with that segment's clip attached.
This is the call that produces the transcript, on-screen text, and visual
description held by a segment.

\begin{pbox}
\begin{lstlisting}[style=promptbox]
Analyse the video segment from {start}s to {end}s.
Reply ONLY with a JSON object:
{"transcript":"<speech, transcribed from the audio>",
 "ocr_text":"<text visible on screen>",
 "visual_description":"<what is visible>",
 "summary":"<a concise summary>",
 "tags":["tag1","tag2"],
 "key_moments":[{"time":0.0,"description":"..."}]}
\end{lstlisting}
\end{pbox}

\noindent
We supply no subtitle files, so \texttt{transcript} is always produced by
the backbone from the embedded audio; this is the without-subtitles
protocol stated in the implementation details.

\subsection{Growth Phase}

\noindent\textbf{P3. Controller and sufficiency check.} Issued once per
drill round. It receives the question, every level-$1$ node in temporal
order, and the evidence gathered for the current question, and it returns
the tuple $(\delta,\mathcal{A},\mathcal{X},\mathcal{P})$.

\begin{pbox}
\begin{lstlisting}[style=promptbox]
You are the controller of a hierarchical video
memory. Decide whether the evidence collected so far
already answers the question, and if it does not,
decide which branch to refine next.

Question: {question}

Level-1 event nodes, in temporal order (together
they cover the whole video):
{node_list}

Evidence collected for this question so far:
{evidence}

Reply ONLY with a JSON object:
{"sufficient": true_or_false,
 "read_node_ids": ["E3", ...],
 "expand_node_ids": ["E3", ...],
 "read_plan": {"summary": true_or_false,
               "text_evidence": true_or_false,
               "frames": 0},
 "reasoning": "one sentence"}

Rules:
- sufficient: true ONLY if the collected evidence
  already determines the answer.
- read_node_ids: branches whose stored evidence
  should be read this round.
- expand_node_ids: ordered by priority, used only
  when sufficient is false. The first entry is the
  branch refined this round.
- read_plan: what to retrieve from the branches in
  read_node_ids, that is, node summaries, denser
  textual evidence, or a small number of frames.
- Answer options are not available at this stage. Do
  not guess an option; judge only whether the
  evidence pins the answer down.
- For questions that turn on visible detail such as
  colour, count, or on-screen text, require explicit
  evidence before reporting sufficient.
- For questions anchored at the beginning or at the
  end of the video, prefer branches covering the
  earliest or the latest timestamps respectively.
\end{lstlisting}
\end{pbox}

\noindent
Every level-$1$ node appears in \texttt{\{node\_list\}}: there is no
embedding-based pre-filter and no node-count cutoff, which is what makes any
part of the timeline reachable in a single round. One line per node,
in temporal order:

\begin{pbox}
\begin{lstlisting}[style=tracebox]
[{id}] {start}s-{end}s [{status}]: {summary}
\end{lstlisting}
\end{pbox}

\noindent
\texttt{\{status\}} is \texttt{expanded} or \texttt{unexpanded}, and
\texttt{\{summary\}} is the current level-$1$ summary, which already
reflects the bottom-up updates written by earlier questions. Deeper nodes
are never listed here; they are reached by the local traversal of P4. The
sequence of decisions $\delta_0,\delta_1,\dots$ returned across rounds
defines the stopping rule of Algorithm~\ref{alg:query},
\begin{equation}
r^{\star}=\min\{\,r\le R:\delta_r=1\,\},
\qquad
\rho(q)=\min(r^{\star},R),
\label{eq:stopping}
\end{equation}
with $\min\emptyset=\infty$, so $\rho(q)$ rounds are spent on $q$ and the
budget $R$ binds only when the gate never fires.

\noindent\textbf{P4. In-branch descent.} Once a level-$1$ branch
$b\in\mathcal{X}$ is selected, VTree walks down it one level at a time,
reusing descendants that earlier questions created. Each step scores the
children of the current node against the question and returns the child to
follow, so node selection below level $1$ is made by the model rather than
by a similarity index. The prompt carries no video: it reads only the
stored summaries and textual evidence of the candidate children, which is
what keeps a descent through an already-grown branch free of new
perception.

\begin{pbox}
\begin{lstlisting}[style=promptbox]
You are descending one level of a hierarchical video
memory to find the evidence for a question.

Question: {question}

Current node {node_id} covers {start}s-{end}s.
Its children, in temporal order:
{child_list}

Reply ONLY with a JSON object:
{"stop": true_or_false,
 "next_node_id": "E3.2",
 "sufficient_here": true_or_false,
 "reasoning": "one sentence"}

Rules:
- next_node_id: the child most likely to contain the
  answer. Choose exactly one, by name, from the list.
- sufficient_here: true if the stored summaries and
  evidence of the children already answer the
  question, so that no deeper reading is needed.
- stop: true if no child is a better place to look
  than the current node, for example when the answer
  spans several children or none of them is relevant.
- Judge only from the text shown. Do not assume
  content that is not in a summary.
\end{lstlisting}
\end{pbox}

\noindent
\texttt{\{child\_list\}} uses the same line template as
\texttt{\{node\_list\}} above. The descent stops when \texttt{stop} or
\texttt{sufficient\_here} is returned, when the depth limit is met, or when
it reaches a node with no children; only the last case can trigger P5.
Writing $\mathcal{L}_D(\mathcal{T})=\{n\in\mathcal{L}(\mathcal{T}):
\ell(n)<D\}$ for the expandable leaves, a descent that ends at
$x\in\mathcal{L}_D(\mathcal{T})$ issues one P5 call, and a descent that
ends anywhere else returns its collected evidence to the controller
without any new perception. This is the mechanism by which a branch grown
for question $i$ is reused, rather than regrown, by question $i{+}1$.

\noindent\textbf{P5. Query-conditioned branch expansion.} Issued when the
descent reaches an unexpanded leaf $x\in\mathcal{L}_D(\mathcal{T})$, with
the clip of $x$ attached at a higher visual density than the build phase
used. This prompt is what makes the growth question-shaped rather than a
fixed pyramid: the sub-intervals a branch unfolds into are conditioned on
\texttt{\{question\}}.

\begin{pbox}
\begin{lstlisting}[style=promptbox]
Focus on the video segment from {start}s to {end}s.
Question: {question}
Already known about this segment: {node_context}

Split this segment into at most {max_children}
consecutive subevents, chosen so that the part of
the segment relevant to the question becomes easier
to locate. The subevents must tile the whole range:
the first starts at {start}, the last ends at {end},
and each one begins exactly where the previous one
ends. Each subevent must last at least
{min_duration}s unless the range itself is shorter.
Each summary should say what happens in that
subevent and how it bears on the question.
Reply ONLY with a JSON array:
[{"start":0.0,"end":10.0,"summary":"..."},...]
\end{lstlisting}
\end{pbox}

\noindent
\texttt{\{max\_children\}}$=20$ and \texttt{\{min\_duration\}}$=3$\,s when
the parent span permits it. \texttt{\{node\_context\}} is the stored
summary and textual evidence of $x$. The returned intervals are snapped by
Equation~\ref{eq:normalize} with $[0,T)$ replaced by $[s_x,e_x)$, so they
satisfy Equation~\ref{eq:partition} regardless of what the model returns.
Two consequences are worth stating because they are what the persistence
argument rests on. First, the conditioning enters only through the
\emph{placement} of the boundaries and the wording of the summaries, never
through the coverage: the children still tile the parent, so a later
question may refine a region that was irrelevant to the question which
triggered the expansion. Second, expansion is monotone. Nodes are only
added, never merged or deleted, so the state written back after question
$i$ is a superset of the state read at question $i$, which is why
$\mathcal{T}$ can be reloaded across the question stream without any
per-question reconciliation.

\subsection{Answer Generation}

\noindent\textbf{P6. Answer generation.} Issued once per question, after
the loop terminates. This is the only stage at which multiple-choice
options enter the pipeline.

\begin{pbox}
\begin{lstlisting}[style=promptbox]
You are a video question-answering assistant. You
are given evidence extracted from a video:
timestamps, summaries, transcripts, on-screen text,
and visual descriptions. Keyframes are attached when
the reading plan asked for them.

Evidence:
{evidence_context}

Question: {question}

Options:
{options}

Which option best answers the question?
IMPORTANT: your entire response must be a single
digit: {valid_digits}. Do not write any explanation,
letter, or other text.
\end{lstlisting}
\end{pbox}

\noindent
\texttt{\{options\}} is rendered as \texttt{0: ...}, \texttt{1: ...}, one
per line, and \texttt{\{valid\_digits\}} lists the admissible indices. The
single-token constraint is what the reported accuracies are scored against.
For open-ended questions the last two lines are replaced by
\texttt{Answer concisely, and cite the timestamps you used.}

\subsection{Baseline Prompt}

\noindent\textbf{B1. Direct backbone.} The backbone rows of
Table~\ref{tab:main} are produced by uploading the whole video and asking
the same model to answer directly.

\begin{pbox}
\begin{lstlisting}[style=promptbox]
You are answering a multiple-choice question about
the video above.

Question: {question}

Options:
{options}

Watch the whole video and choose the single best
option.
IMPORTANT: respond with ONLY one digit
({valid_digits}). No letters, no words, no
explanation.
\end{lstlisting}
\end{pbox}

\noindent
The baseline is not handicapped relative to VTree: it uses the same model
version string, the same decoding temperature, the same option rendering,
and receives the entire video rather than a selected subset of it. The
difference between the two rows of each block in Table~\ref{tab:main} is
therefore attributable to what is read, not to how it is asked.

\subsection{Settings and Truncation}

Table~\ref{tab:prompt_params} collects the settings that a reproducer needs
in addition to the prompt text. Every prompt that carries video is sent at
low media resolution, including the baseline.

\begin{table}[h]
\centering
\small
\setlength{\tabcolsep}{4pt}
\begin{tabular}{@{}llc@{}}
\toprule
Setting & Applies to & Value \\
\midrule
Frame decoding rate      & Build             & $1$ fps \\
Max children             & P5                & $20$ \\
Min child duration       & P5                & $3$\,s \\
Frames stored per node   & P5                & $3$ \\
Max tree depth $D$       & Growth            & $4$ \\
Drill budget $R$         & Growth            & $5$ \\
\midrule
Temperature              & P1, P2, P3, P4, P5 & $0.1$ \\
Temperature              & P6, B1            & $0.0$ \\
Media resolution         & video calls       & low \\
\bottomrule
\end{tabular}
\caption{Decoding and structural settings for the prompts of
Section~\ref{sec:appendix_prompts}.}
\label{tab:prompt_params}
\end{table}

Every field that a prompt interpolates is length-capped, so that a long
video cannot push a prompt past the context window of the backbone and so
that prompt length does not drift with the size of the accumulated tree.
Table~\ref{tab:truncation} lists the caps. The one worth attention is the
$120$-character cap on a node summary inside \texttt{\{node\_list\}}: the
level-$1$ list is never truncated by node count, since global reach depends
on every branch being visible, so the length of the routing prompt is held
down per line instead. A branch that has been expanded many times therefore
contributes a longer summary to its own subtree but the same budget to the
routing view.

\begin{table}[h]
\centering
\small
\setlength{\tabcolsep}{4pt}
\begin{tabular}{@{}llc@{}}
\toprule
Field & Prompt & Cap \\
\midrule
\texttt{\{question\}}          & P3, P4     & $600$ ch. \\
\texttt{\{evidence\}}          & P3         & $2000$ ch. \\
Node summary, per line         & P3, P4     & $120$ ch. \\
Candidate nodes                & P3         & uncapped \\
\texttt{\{node\_context\}}     & P5         & $300$ ch. \\
\texttt{\{evidence\_context\}} & P6         & $8000$ ch. \\
Evidence text, per line        & P3, P6     & $300$ ch. \\
Parent summary after update    & write-back & $300$ ch. \\
\bottomrule
\end{tabular}
\caption{Truncation applied to interpolated fields. Truncation is by
characters and is applied after the ordering and deduplication of
Section~\ref{sec:appendix_prompts}, so the earliest evidence survives.}
\label{tab:truncation}
\end{table}

\subsection{Text Templates (No Model Call)}

Four string templates shape what the model sees but involve no model call.
A reviewer should not mistake them for prompts.

\noindent\textbf{Evidence rendering.} The value bound to
\texttt{\{evidence\}} in P3 and to \texttt{\{evidence\_context\}} in P6 is
one line per evidence item:

\begin{pbox}
\begin{lstlisting}[style=tracebox]
[{start}s-{end}s] [{type}] {text}
\end{lstlisting}
\end{pbox}

\noindent
\texttt{\{type\}} is one of \texttt{summary}, \texttt{transcript},
\texttt{ocr}, or \texttt{visual}. Because the same region may be read again
in a later round, or by a later question that reuses the branch, items are
keyed by
\begin{equation}
\kappa(e)=\bigl(\tau(e),\ \lfloor 2s_e\rfloor,\ \lfloor 2e_e\rfloor\bigr),
\label{eq:dedup}
\end{equation}
that is by type together with the span quantized into half-second buckets.
One item is kept per distinct key and the survivors are ordered by $s_e$,
so the rendered block is a time-ordered reading of the video with no
repeated spans. Paths of the retrieved frames are appended after the text
lines, and the field is \texttt{(no evidence)} on the first round, before
anything has been read.

\noindent\textbf{Representative frames.} The frames attached to P6, and the
three frames stored on each new node created by P5, are drawn from the
cached $1$\,fps decode of the node span. No model call and no learned
selector is involved; the count requested by the controller through
\texttt{read\_plan} is the only query-dependent quantity.

\noindent\textbf{Global video summary.} The build phase concatenates the
level-$1$ summaries in temporal order under a fixed length limit. It is a
concatenation, not a further model call.

\noindent\textbf{Parent summary update.} After an expansion, the enriched
parent summary is assembled as

\begin{pbox}
\begin{lstlisting}[style=tracebox]
{parent_summary}. Details: {child_summaries}
\end{lstlisting}
\end{pbox}

\noindent
where \texttt{\{child\_summaries\}} joins the summaries of the new children
in temporal order under a bounded total length, and the result is
propagated recursively to the level-$1$ ancestor. Ancestor summaries are
therefore assembled from their descendants rather than regenerated by the
model, which is why the bottom-up update of Section~\ref{sec:method} costs
nothing beyond the expansion that triggered it, and why an update can never
contradict the subtree it summarizes.

\subsection{A Worked Exchange}

The following abbreviates the first example of
Figure~\ref{fig:visualization}, showing the JSON actually exchanged. The
question asks which type of raw meat is introduced last. Summaries are
shortened here for space.

\begin{pbox}
\begin{lstlisting}[style=tracebox]
round 0, P3 out:
{"sufficient": false,
 "read_node_ids": ["E12"],
 "expand_node_ids": ["E12"],
 "read_plan": {"summary": true,
               "text_evidence": true, "frames": 0},
 "reasoning": "E12 is the fresh meat market
  segment, but its summary names no meat."}

round 0, P4 out (E12 is a leaf, so P5 fires):
{"stop": false, "next_node_id": null,
 "sufficient_here": false,
 "reasoning": "E12 has no children yet."}

round 0, P5 out:
[{"start":..., "end":..., "summary":"eating in the
  night market"},
 {"start":..., "end":..., "summary":"a rare wild raw
  meat is introduced"},
 {"start":..., "end":..., "summary":"night market
  preparation and eating"}]

round 1, P3 out:
{"sufficient": false, "read_node_ids": ["E12"],
 "expand_node_ids": ["E12"], "read_plan": {...},
 "reasoning": "the child summaries name a meat
  introduction but not the meat."}

round 1, P4 out:
{"stop": false, "next_node_id": "E12.2",
 "sufficient_here": false,
 "reasoning": "E12.2 is the introduction itself."}

round 1, P5 out (expanding E12.2):
[{... "summary":"the head chef debones the deer"},
 {... "summary":"raw deer meat is introduced"},
 {... "summary":"the hosts taste the meat"}]

round 2, P3 out:
{"sufficient": true, "read_node_ids": ["E12"],
 "expand_node_ids": [],
 "read_plan": {"summary": true,
               "text_evidence": true, "frames": 3},
 "reasoning": "the last introduced raw meat is
  named, with frames available."}

P6 out:
1
\end{lstlisting}
\end{pbox}

\noindent
Two rounds of growth were spent, against a budget of five, and the tree
gained five nodes that any later question about this video inherits. The
exchange also shows the interface separation the Method relies on: no
option text appears anywhere before the final call, so the branch that was
descended was chosen on the question stem alone.

\end{document}